# Accurate Merging of Images for Predictive Analysis using Combined Image

T.R. Gopalakrishnan Nair
Member, Advanced Imaging and Computer Vision Group
Aramco Endowed Chair, PMU, KSA
VP, RIIC, Dayananda Sagar Institutions
Bangalore, India
trgnair@gmail.com

Richa Sharma
Member, Advanced Imaging and Computer Vision Group
Research and Industry Centre (RIIC), Dayananda Sagar Institutions
Bangalore, India
richarichas@yahoo.com

*Abstract*— Several Scientific and engineering applications require merging of sampled images for complex perception development. In most cases, for such requirements, images are merged at intensity level. Even though it gives fairly good perception of combined scenario of objects and scenes, it is found that they are not sufficient enough to analyze certain engineering cases. The main problem is incoherent modulation of intensity arising out of phase properties being lost. In order to compensate these losses, combined phase and amplitude merge is demanded. We present here a method which could be used in precision engineering and biological applications where more precise prediction is required of a combined phenomenon. When pixels are added, its original property is lost but accurate merging of intended pixels can be achieved in high quality using frequency domain properties of an image. This paper introduces a technique to merge various images which can be used as a simple but effective technique for overlapped view of a set of images and producing reduced dataset for review purposes.

## I. INTRODUCTION

There are several scenarios in image engineering where cases exist for multiple object merging under common denominator scenes and later separating these multi objects present in distinctively in intensity separable layers. It could be scenarios of night vision for security processing or it could be microscopic vision of biological specimens and culture. Though fundamentally, Images are a collection of pixels and each pixel is represented by some values depending on the type of the image, dealing them at intensity level is insufficient for precision management. In gray level digital images, pixels contain intensity information and we perceive and analyse an image based on the changes in shades of color intensities or frequencies [1].

Edges are formed due to sudden change in intensity and we can visualize and identify various objects in an image because of their color, shapes and texture. It is the rate of change of intensity values which gives the illusion of an object. In the field of image processing, color/ gray level intensity information can be processed easily as the image is stored as a collection of pixels of various colors or intensities.

Each object in an image generates a unique spectrum of frequencies in frequency domain. An image with many such objects contains a collection of all those frequencies. This paper demonstrates the feasibility of creating accurate high-fidelity images through merging, for prediction purposes in biology and solid mechanics. In order to cater to the needs of complete information of image, it is necessary and better to merge the frequencies along with phase to get overlapped view of the objects or images. Such methods are required for several practices in astronomy, radiology, body implant predictions, fracture mechanics prediction etc.

In literature, Image merging has been used for many reasons like improving resolution or implementing segmentation. A novel image restoration algorithm to deblur the image without estimating the image blur by merging differently blurred multiple images in the spectrum domain using the fuzzy projection onto convex sets (POCS) can be found in [2]. Statistical Region Merging (SRM) and the Minimum Heterogeneity Rule (MHR) have also been used for object merging [3]. The SRM segmentation method not only considers spectral, shape, and scale information, but also has the ability to cope with significant noise corruption and handle occlusions. The MHR used for merging objects takes advantage of its spectral, shape, scale information, and the local and global information. A novel self-adaptive weighted average fusion scheme based on standard deviation of measurements to merge IR and visible images is developed in the special domain using the better recovery tool of total variation optimization [4]. It achieves a high level of fusion quality in global information.

Also, in computer vision, multisensor image fusion is the process of combining relevant information from two or more images into a single image. The resulting image will be more informative than any of the input images. Image fusion has become a common term used within medical diagnostics and treatment too. The term is used when certain portions of multiple images of a patient are registered and overlaid or merged to provide additional combined information. Fused images may be created from multiple images from the same imaging modality [5], or by combining information from multiple modalities [6], such as magnetic resonance image (MRI), computed tomography (CT), positron emission tomography (PET), and single photon emission computed tomography (SPECT). In radiology and radiation oncology, these images serve different purposes. For example, CT images are used more often to ascertain differences in tissue density while MRI images are typically used to diagnose brain tumors. Merging prostate imaging has also been used to identify the location and aggressiveness of prostate cancer [7].

Usually these practices are giving importance to intensity management rather than phase coordination in merging.

In remote sensing applications, the increasing availability of space borne sensors gives a motivation for different image fusion algorithms. Several situations in image processing require high spatial and high spectral resolution in a single image. Most of the available equipments are not capable of providing such data convincingly [12]. The image fusion techniques allow the integration of different information sources. The fused image can have complementary spatial and spectral resolution characteristics. However, the standard image fusion techniques can distort the spectral information of the multispectral data while merging. Data fusion method for land cover (LC) classification that combines remote sensing data at a fine and a coarse spatial resolution can be found in [8]. This classifier uses all image information (bands) available at both fine and coarse spatial resolutions by stacking the individual image bands into a multidimensional vector.

Usually merging of different data sets is used in digital image processing to improve the visual and analytical quality of the data [9]. The analyst may need to merge different types of data. In this process, different data such as satellite imagery from the same sensor but with different resolution, satellite imagery from different sensors with varying resolution, digitized aerial photography and satellite imagery or satellite imagery with ancillary information can be merged. There are many techniques for merging like Principal Component, IHS, and Brovey Transform. A technique for multi-image fusion in one-pass through overlapping input images which restores and reconstructs the scene radiance field can be found in [10]. The technique is effective because it maximizes fidelity based on a comprehensive end-to-end system model that accounts for scene statistics, acquisition blurring, sampling, and noise

This paper is not about resolution merge, it is about object or images merge. Merging of images discussed in this paper can be used in many applications for example, solid property predictions, creating currency, body implant, compression fractures, fast viewing or creative editing. In biological systems multi cellular growth patterns created separately can be viewed as single slide for studies of growth which is required for damage characteristics prediction. The paper demonstrates merging in spatial domain as well as in frequency domain and it demonstrates that merging in frequency domain gives better performance as it preserves key properties of original images in a better way.

## II. SPECTRAL TREATMENT

The repetitive nature or the frequency characteristics of images can be analyzed using spectral decomposition methods like Fourier analysis. The precision of fusion much depends on the phase properties of pixels in streams rather than the individual intensity property which can yield only a coarse merge. In an RXC (*Row × Column*) digital image, positions *u* and *v* indicate the number of repetitions of the sinusoid in those directions. Therefore the wavelengths along the column and row axes are

$$\lambda_u = \frac{C}{u} \quad \text{and} \quad \lambda_v = \frac{R}{v} \quad \text{pixels,}$$

and the wavelength in the wavefront direction is

$$\lambda_{wf} = \sqrt{\left(\frac{C}{u}\right)^2 + \left(\frac{R}{v}\right)^2}.$$

The frequency is the fraction of the sinusoid traversed over one pixel,

$$\omega_u = \frac{u}{C}, \quad \omega_v = \frac{v}{R}, \text{ and}$$

$$\omega_{wf} = 1 \Big/ \sqrt{\left(\frac{C}{u}\right)^2 + \left(\frac{R}{v}\right)^2} \quad \text{cycles.}$$

The wavefront direction is given by

$$\theta_{wf} = \tan^{-1}\left(\frac{\omega_v}{\omega_u}\right) = \tan^{-1}\left(\frac{vC}{uR}\right).$$

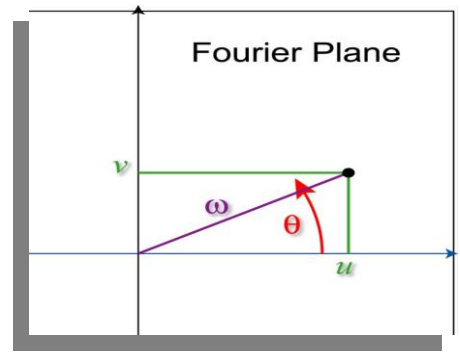

Fig. 1. Details of Fourier Plane

It is well known that edges in an image are generated by high frequencies and human vision system is more sensitive to edges as compared to constant or slow varying intensities [11] and high frequency coefficients tend to be very small and they can be quantized very effectively without distorting the results to achieve data reduction [12]. This paper also exploits this fact for data reduction of the images after merging by utilising a threshold T in algorithm B of section 3.

## III. IMPLEMENTATION

This section describes the algorithms used for precision merging. We have created a prototype implementation for object merging using the following algorithms:

*Merging in spatial domain:*

A. Algorithm

Initialization:

Set the value for number of images/layers 'n'
Align the images or layers so that number of samples is same.
Normalize the image values i(x,y) to 0-1
**While** all pixels(x,y) in the image are not seen **do**
ADD corresponding intensities of all n images

$$Result(x, y) = i_1(x, y) + ...... + i_n(x, y)$$

*Merging in frequency domain:*

Linear integral of frequency density from different images like in different windows of FFT, is a promising approach to create better accuracy. In the total spectral spread, the images differ by spectral density per window. This spectral strength corresponding to each image at corresponding windows is independent of intensity. We make use of this fact in the following algorithm and its unique success is shown in the following figures.

Suppose there are 'n' images to be merged each of size RXC (*Row × Column*). In frequency spectrum of an RXC digital image, positions *u* and *v* indicate the number of repetitions of the sinusoid in those directions. We scale each sinusoid with a *prominence coefficient* before integration for perception control. As per definition of Fourier Transform, frequency spectrum of image '*i*' can be represented by the following formulae:

$$FFT(i(x,y)) = I(u,v) = \sum_x \sum_y i(x,y) \exp[-j2\pi(ux/R + vy/C)] \quad (1)$$

Hence,

$$I(u,v) = \sum_x \sum_y i(x,y)[\cos(\theta) - j\sin(\theta)] \quad (2)$$

Where,

$$\theta = 2\pi(ux/R + vy/C) \quad (3)$$

Same can be rewritten as:

$$I(u,v) = \sum_x \sum_y i(x,y)[R_\theta - I_\theta] \quad (4)$$

Where, *R* and *I* stand for real part and imaginary part of Fourier Spectrum.

Suppose there are n images to be merged and each image pixel p can be represented as $p_{xyn}$ in 3D domain of n images and its corresponding frequency $P_{uvn}$ can be calculated by the following equation:

$$P_{uvn} = a_n \sum_x \sum_y i_n(x,y)[R_\theta - I_\theta]$$

And

$$P_{integral} = \sum_n P_{uvn}$$

Where $a_1, a_2....a_n$ are prominence coefficients.
Therefore,

$$P_{integral} = a_1 \sum_x \sum_y i_{1(x,y)}[R_\theta - I_\theta] + .....$$

$$..... + a_n \sum_x \sum_y i_{n(x,y)}[R_\theta - I_\theta]$$

Actual values of these *prominence coefficients* will be adjusted by visual perception according to the application and requirement. Further study of perception coefficients is beyond scope of this paper.

B. Algorithm

Initialization:

Set the value for number of images/layers 'n'
Align the images or layers so that number of samples is same.
Normalize the image values i(x,y) to 0-1
Identify the highest frequency (Max.Frequency) from frequency spectrum of the input image.
Set threshold T=Max.Frequency * x
Where 0<x<1
**While** all pixels(x,y) in the images are not seen **do**
ADD corresponding density of frequencies of all n images

$$Result(u,v) = a_1 FFT(i_1(x,y)) + ...... + a_n FFT(i_n(x,y))$$

Remove all frequencies below threshold 'T' to achieve data reduction.
Take inverse FFT to visualize the merged image.

## IV. RESULT AND ANALYSIS

The experiments were aimed at developing a system to merge images in such a way that all images can be seen together in high fidelity without much loss of information because of the overriding intensity factors.

The comparative study was performed over ten images for merging in spatial domain as well as in frequency domain by using Algorithm A and Algorithm B as discussed in section 3. It is found by visual inspection that merging in spatial domain as well as frequency domain both gives similar results if the objects in input images are spatially separate or non overlapping from each other (Fig.2) whereas merging in frequency domain out performs if objects in the images are getting overlapped in merged image. We have not gone for quantitative methods to further check for error occurred while imaging as visual inspection itself gave satisfactory evidence. Merging in frequency domain is capable of keeping fine details of all input images (Fig. 3). This technique can be useful in surveillance where the observer is supposed to see multiple pictures coming from many cameras at a time as it gives ease to the observer to view multiple images at a time in a single screen as well as with reduce data load. If we merge more than 2 images, we can get further compressed view. For example, five different images can be merged and viewed together as one image (Fig.2).

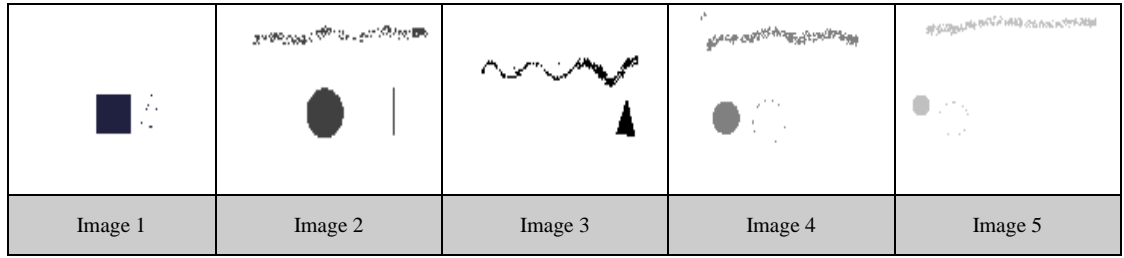

(a)

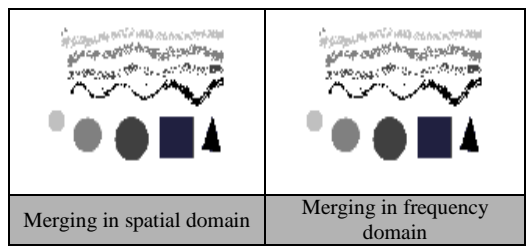

(b)

Fig. 2. (a). Artificially generated Input Images (b). Merged image generated by Algorithm A and B for n=5. Output quality looks same in both cases.

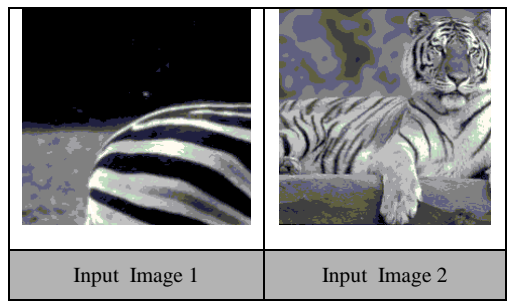

(a)

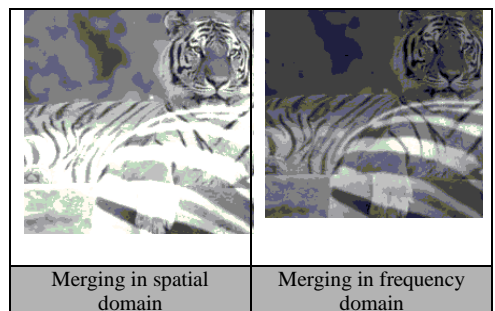

(b)

Fig. 3. (a).Natural Input Images to be merged (b).Merging using Algorithm A and Algorithm B for n=2. Tiger's paw is more clearly seen in second image.

Merging itself is a form of data reduction and it is possible to further reduce the data required to represent the merged images by retaining frequencies with higher coefficient values. This feature has been implemented by providing a threshold T in algorithm B of section 3. High frequency coefficients tend to be very small and they can be quantized very effectively without distorting the results to achieve compression [12]. The effect of data reduction achieved by us on merged images can be seen in Fig. 4. A reduction ration of 10 (10:1) means that the first data set has 10 information carrying units for every 1 unit in the second (reduced) data set. An attempt has been made to show that it is not required to preserve all frequencies to visualize the merged image; the best reduction ratio for a specific application depends on the application and can be chosen iteratively as more reduction comes with more loss of information. The specific outputs of Fig. 4 demonstrate that in this case, reduction ratio of 8 can be achieved without any visible loss whereas reduction ratio of 22 shows significant loss as body of the tiger got merged with the background.

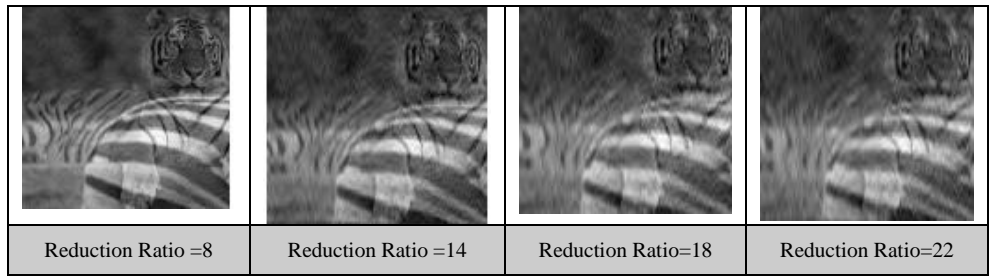

| Reduction Ratio =8 | Reduction Ratio =14 | Reduction Ratio=18 | Reduction Ratio=22 |

Fig. 4. Merging with reduced dataset: Reduction ratio of 8 can be achieved easily without any visible loss. Image with Reduction ratio of 22 shows significant loss as body of the tiger got merged with the background.

## V. CONCLUSION

This work has developed a simple but effective technique for object merging using linear integration of spectral density in corresponding windows. The images merged in this way has displayed improved precision, which is capable of being used in different application like fracture prediction in solids, multicellular integration in biology and visualization of body implants. This paper demonstrates that the linear integral of frequency density of merging images retains the details of all original images accurately and it can be viewed in a single frame and simultaneously provide a data reduced form. This fact can be widely utilised at many places like, body implant predictions, creating secure currency, fracture mechanics prediction etc.

REFERENCES

[1] Gonzalez, Woods, Eddins, Digital Image Processing using MATLAB, 2nd ed., Gatesmark Publication, 2009.
[2] Seung-Won Jung , Tae-Hyun Kim and Sung-Jea Ko , " A Novel Multiple Image Deblurring Technique Using Fuzzy Projection onto Convex Sets", Signal Processing Letters, IEEE, March 2009.
[3] Haitao Li , Haiyan Gu , Yanshun Han and Jinghui Yang , " An Efficient Multiscale SRMMHR (Statistical Region Merging and Minimum Heterogeneity Rule) Segmentation Method for High-Resolution Remote Sensing Imagery", IEEE J. Applied Earth Observations and Remote Sensing, vol. 2, pp. 67 – 73, June 2009.
[4] Li X., Qin S. Y., "Efficient fusion for infrared and visible images based on compressive sensing principle", IET Image Processing, vol. 5(2), 2011.
[5] Gooding MJ, Rajpoot K, Mitchell S, Chamberlain P, Kennedy SH, Noble JA, "Investigation into the fusion of multiple 4-D fetal echocardiography images to improve image quality", Ultrasound in Medincine and Biology, vol. 36(6), pp. 957-66, 2010.
[6] J.B. Maintz, M.A. Viergever, "A survey of medical image registration", Medical Image Analysis, vol. 2(1), pp. 1-36, 1998.
[7] Huber, J.S., Qiyu Peng, Moses W.W., Reutter B.W., Pouliot J., Hsu I.C., "Development of a PET-Transrectal Ultrasound Prostate Imaging System", IEEE trans. on Nuclear Science, vol. 58(3), 2011.
[8] Kempeneers P., Sedano F., Seebach L., Strobl P., San-Miguel-Ayanz,J. , "Data Fusion of Different Spatial Resolution Remote Sensing Images Applied to Forest-Type Mapping", IEEE Trans. on Geoscience and Remote Sensing, vol. 49(12), 2011.
[9] M. Erdogan, H.H. Maras, A. Yilmaz, Ö.T. Özerbil, "Resolution merge of 1:35.000 scale aerial photographs with landsat 7 ETM imagery", The International Archives of the Photogrammetry, Remote Sensing and Spatial Information Sciences, vol. XXXVII. Part B7, Beijing 2008.
[10] Stephen E. Reichenbach and Jing Li,"Restoration and Reconstruction from Overlapping Images for Multi-Image Fusion", IEEE trans. on Geosciences and Remote Sensing, vol. 39, no. 4, April 2001.
[11] Ralf Steinmetz, Klara Nahrstedt, Media Coding and Content Processing, 2nd ed., Prentice Hall, 2002.

[12] G.Rahul1, R.Kalpana2, "Block-Based Feature Multi Level Multi Focus Image Fusion on Neural Network", Int. J. of Engineering Research & Technology (IJERT), vol. 1(8), October 2012.